\documentclass[10pt,twocolumn,letterpaper]{article}

\usepackage{iccv}
\usepackage{times}
\usepackage{epsfig}
\usepackage{graphicx}
\usepackage{amsmath}
\usepackage{amssymb}
\usepackage{comment}
\usepackage{float}
\usepackage{pagecolor}
\usepackage{xcolor}

\pdfpxdimen=\dimexpr 1 in/72\relax

\newcommand\vcr{V \mapsto C \times R}
\newcommand\crv{C \times R \mapsto V}

\newcommand\bo[1]{\textbf{#1}}

\newcommand\A{\textbf{A}}
\newcommand\AV{\textbf{A}_V}
\newcommand\AC{\textbf{A}_C}
\newcommand\D{\textbf{D}}
\newcommand\E{\textbf{E}}

\newcommand{\eq}[1]{\begin{equation}{#1}\end{equation}}

\newcommand{\norm}[1]{\left\lVert#1\right\rVert}

\usepackage{booktabs}
\usepackage{subfig}
\usepackage{comment}

\usepackage[pagebackref=true,breaklinks=true,letterpaper=true,colorlinks,bookmarks=false]{hyperref}

\iccvfinalcopy 

\def\iccvPaperID{23} 

\ificcvfinal\fi
\begin{document}

\newcommand\katef[1]{\textcolor{red}{#1}}



\title{Image Disentanglement and Uncooperative Re-Entanglement \\for High-Fidelity Image-to-Image Translation}

 
 
\author{Adam W. Harley \thanks{Work done while at Facebook Reality Labs.} \\
Carnegie Mellon University\\
{\tt\small aharley@cmu.edu}
\and
Shih-En Wei\\
Facebook Reality Labs\\
{\tt\small shih-en.wei@oculus.com}
\and
Jason Saragih\\
Facebook Reality Labs\\
{\tt\small jason.saragih@oculus.com}
\and
Katerina Fragkiadaki\\
Carnegie Mellon University\\
{\tt\small katef@cs.cmu.edu}
}
\maketitle


\begin{abstract}

Cross-domain image-to-image translation should satisfy two requirements: (1) preserve the information that is common to both domains, and (2) generate convincing images covering variations that appear in the target domain. This is challenging, especially when there are no example translations available as supervision. Adversarial cycle consistency was recently proposed as a solution \cite{zhu2017unpaired}, with beautiful and creative results, yielding much follow-up work. However, augmented reality applications cannot readily use such techniques to provide users with compelling translations of real scenes, because the translations do not have high-fidelity constraints. In other words, current models are liable to change details that should be preserved: while re-texturing a face, they may alter the face's expression in an unpredictable way. In this paper, we introduce the problem of high-fidelity image-to-image translation, and present a method for solving it. Our main insight is that low-fidelity translations typically escape a cycle-consistency penalty, because the back-translator learns to compensate for the forward-translator's errors. We therefore introduce an optimization technique that prevents the networks from cooperating:  simply train each network only when its input data is real. Prior works, in comparison, train each network with a mix of real and generated data. Experimental results show that our method accurately disentangles the factors that separate the domains, and converges to semantics-preserving translations that prior methods miss. 
\vspace{-1.5em}
\end{abstract}

\begin{figure}[t!]
\centering
\includegraphics[width=1.0\linewidth]{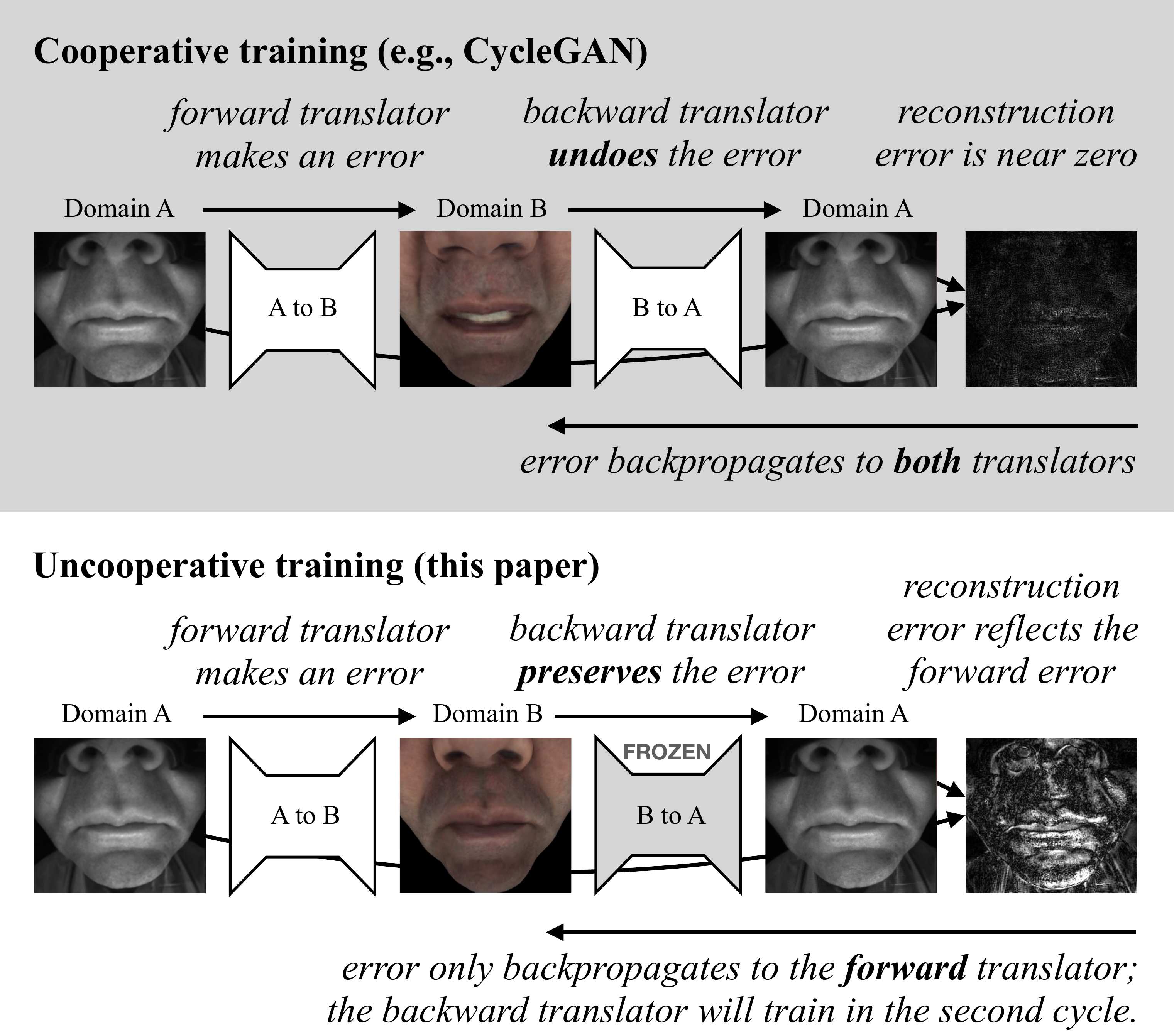}
\caption{``Coooperative'' training setup from prior work (top), compared with our proposed ``uncooperative'' training (bottom). Preventing cooperation between the forward and back-translation yields a more useful reconstruction loss.
}
\label{fig:teaser}
\vspace{-1em}
\end{figure}

\section{Introduction}\label{sec:intro}

Unpaired cross-domain image-to-image translation is achieving exceptionally convincing results in a variety of domains \cite{zhu2017unpaired}. 
\textit{High-fidelity} image translation requires not only credible translations, but also strict preservation of the factors that are common to both domains. 
Consider Figure~\ref{fig:teaser}. We wish to translate an image of a face across two domains that mostly differ in texture. It is inappropriate for the translator to additionally change the face's expression. Unfortunately, this failure mode is surprisingly common in standard unsupervised image-to-image models. 

Why does this happen? In the standard approach (based on CycleGAN \cite{zhu2017unpaired}), there are two main networks: a forward translator, and a backward translator. There are also two main losses: an adversarial loss, which encourages the translated images to be indistinguishable from ones in their target domain, and a cycle consistency loss, which encourages that forward translation (\ie, A to B) followed by backward translation (\ie, B back to A) yields the original input (\ie, forming a cycle). The problem with this setup is that there is no ``fidelity loss'' on the translation. In other words, the forward translator may generate arbitrary samples in domain B, and as long as the backward translator reconstructs the input, there is no penalty. One can add geometric constraints to the translation \cite{gcgan, mueller2018ganerated}, but these only approximate a loss on fidelity. 
We present an approach that directly penalizes unfaithful translations, by ensuring that forward-translation errors are \textit{preserved} during back-translation, as shown in Figure~\ref{fig:teaser} (bottom). 

\begin{figure}[t!]
\centering
\includegraphics[width=1.0\linewidth]{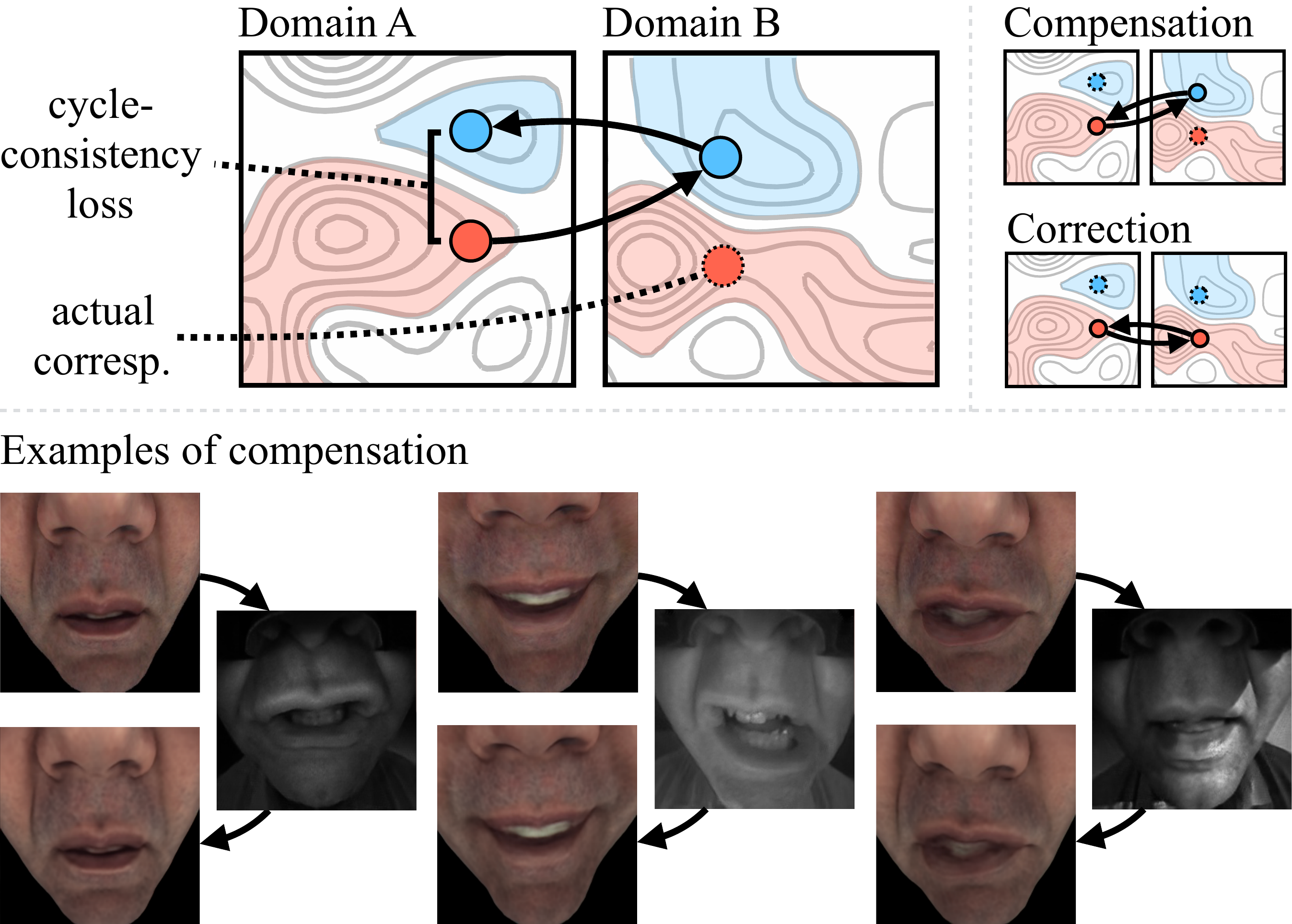}
\caption{
Top: Two strategies for optimizing the cycle-consistency loss: compensation and correction. 
Optimizing the cycle-consistency loss 
encourages the forward-translator to find better mappings (i.e., correction), but also encourages the back-translator to compensate for bad mappings (i.e., compensation). 
Bottom: real examples of compensation learned by a CycleGAN; cycle-consistency is being met, but the translations are not faithful. 
}
\label{fig:cycleloss}
\vspace{-1.5em}
\end{figure}

Our main insight is that cycles are problematic when the two mapping functions are allowed to \textit{cooperate}. By ``cooperate'' we mean that they optimize for each other's outputs. 
CycleGAN and its many variants \cite{zhu2017unpaired,huang2018multimodal,lee2018diverse,gonzalez2018image,liu2017unsupervised,almahairi2018augmented} all have a cooperative training setup: in each cycle, the first translator receives a real input, and the second translator receives a fake input (\ie, an attempted translation/disentanglement) which it back-translates, and \textit{both} networks get penalized according to the reconstruction error. This essentially asks the second network to compensate for the first network's errors. This is counter-productive, because if the second network succeeds, then the first network need not improve. Given sufficient optimization time, these cooperative setups find extremely effective ``cheats'', in which subtle signals are encoded into low-fidelity forward translations and subsequently decoded to achieve near-perfect back-translation, thus defeating the reconstruction error \cite{chu2017cyclegan}. This is illustrated in Figure~\ref{fig:cycleloss}. 

Our main contribution is in preventing the networks from compensating for each other's errors, via a simple optimization technique: simply train each network \textit{only} when its input data is real. With this technique, neither network learns about the other's behavior, which renders cooperation impossible. Instead, the back-translator simply \textit{preserves} any errors made during forward-translation, and the reconstruction penalty is put entirely on the forward translator. 
This forces the networks to learn more faithful mappings to their target domains. 

Our ``uncooperative'' training also provides a route to unsupervised factor disentanglement. 
Several prior works have modified CycleGANs to perform a disentanglement and subsequent re-entanglement: the first translator disentangles the input into (1) an image in the second domain, and (2) a residual; the second translator entangles these to reconstruct the input \cite{zhu2017toward,huang2018multimodal,almahairi2018augmented}. In practice, however, 
an unconstrained ``residual'' path can actually be detrimental to the final results, 
since the model may exploit this path to encode the \textit{entire input,} greatly facilitating the the cycle-consistency objective. Prior works have proposed a variety of methods to mitigate this problem, but usually at the cost of severely reducing the representational capacity of the residual (\eg, limiting it to 8 dimensions), and making assumptions about its distribution (\eg, assuming it is standard normal) \cite{almahairi2018augmented,huang2018multimodal,lee2018diverse,liu2017unsupervised}. After applying these constraints, some prior works report that the residual path simply goes \textit{ignored} by the model, unless its usage is facilitated by careful design choices (\eg, using the residual as layer-wise normalization coefficients) \cite{almahairi2018augmented,liu2017unsupervised}. Our optimization technique allows us to disentangle multi-scale high-dimensional residuals, without requiring parameter-sensitive representational constraints. 

In experiments with real images, we show that our optimization method delivers an obvious qualitative improvement over the current state-of-the-art, both in terms of semantics-preservation and residual-factor disentanglement. In synthetic data (where the residual is known), we demonstrate that our ``uncooperative'' optimization leads to quantitatively accurate disentanglement, whereas ``cooperative'' optimization does not. 
\section{Related  Work} \label{sec:related}


\textbf{Image-to-image translation} has recently attracted great attention, partly thanks to the success of generative adversarial networks (GANs)~\cite{goodfellow2014generative,mirza2014conditional, aaron16conditional, kossaifi2017gagan}.
The goal in image-to-image translation is to translate an image in one domain to a corresponding image in the second domain. 
%
Pix2Pix~\cite{isola2017image} trains models for this task using paired data from the two domains (\ie, input-output pairs, exemplifying good translations). 
%
CycleGAN~\cite{zhu2017unpaired} removes the need for paired data by forming a translation ``cycle''---forward translation followed by backward translation---which permits a natural reconstruction objective between the input and the back-translation. 
This is important, because in many domains, paired examples do not exist (\eg, a face in the exact same pose/expression in two different physical environments). 
CycleGAN often preserves the structural content of the images, but this may simply be a consequence of the convolutional architecture \cite{lenc2015understanding}.  
%
%
CycleGAN is only capable of learning one-to-one mappings, but several works (not all unsupervised) have proposed variants that are capable of one-to-many mappings, such as Augmented CycleGAN~\cite{almahairi2018augmented}, DRIT~\cite{lee2018diverse}, MUNIT~\cite{huang2018multimodal}, BicycleGAN \cite{zhu2017toward}, and cross-domain disentanglers \cite{gonzalez2018image}. These methods are able to generate diverse image with similar ``content'' (\ie, structural pattern) but different ``style'' (\ie, textural rendering) through disentanglement. 
These methods 
use strong assumptions or regularizations to avoid undesirable local optima, including shared latent spaces \cite{huang2018multimodal,lee2018diverse,gonzalez2018image}, loss on KL divergence from simple Gaussians \cite{huang2018multimodal,lee2018diverse,zhu2017toward}, or low-dimensional representations \cite{lee2018diverse,almahairi2018augmented,zhu2017toward,gonzalez2018image}. The effectiveness of these methods is therefore highly dependent on parameter selection. 

\textbf{Image factor disentanglement} is necessary if we wish to control the latent factors in the generated images.
Hadad \etal~\cite{hadad2018two} assumes the availability of attribute labels in a particular domain, where the goal is to disentangle images into a target domain plus a residual (\ie, ``everything else'').
Many disentanglement works also make strong assumptions on domain
knowledge of the latent space, which includes having data pre-grouped according to individual factors \cite{reed2014learning, kulkarni2015deep}, or having exact knowledge of the structure and function of individual factors (\eg, for faces: identity, pose, shape, texture \cite{shu2017neural, shu2018deforming}).
In this work, we do not have attribute labels, we do not make assumptions on the latent space, and we perform disentanglement using only the unpaired image data. 
Similar to our method, InfoGAN \cite{chen2016infogan} and MINE \cite{belghazi2018mine} are completely unsupervised, but the approach in these works is quite different: these methods maximize the mutual information between the inferred latent variables and the observations, while we use discriminators and reconstruction to achieve disentanglement. 

\section{Method}\label{sec:model}

There are three key ingredients to our method: (1) adversarial priors, which
encourages the translated images to be indistinguishable from ones in their target domain, (2) cycle-consistency, which encourages the translations to be invertible, and (3) ``uncooperative'' optimization, which ensures the networks do not ``cheat'' toward an undesirable local minimum.

\subsection{Preliminaries}


Let $V$ and $C$ be two image domains, such that the images $v \in V$ have more information than the images $c \in C$. That is, $V$ contains variation in some latent factor that is either constant or absent in $C$. This implies that the mapping $V \mapsto C$ is many-to-one, and the mapping $C \mapsto V$ is one-to-many. As a mnemonic, note that $V$ is \textbf{v}ariable in some aspect where $C$ is \textbf{c}onstant.

Let $R$ be the \textbf{r}esidual information that is in $V$ but not in $C$. Accessing this extra information allows us to form bijective (one-to-one) mappings, $\vcr$ and $\crv$. Note that $R$ is not necessarily an image domain. In our implementation, each $r \in R$ is a collection of deep featuremaps at multiple scales, which allows its actual form to be determined entirely by the data. 

\begin{figure*}[t!]
\centering
\includegraphics[width=1.0\linewidth]{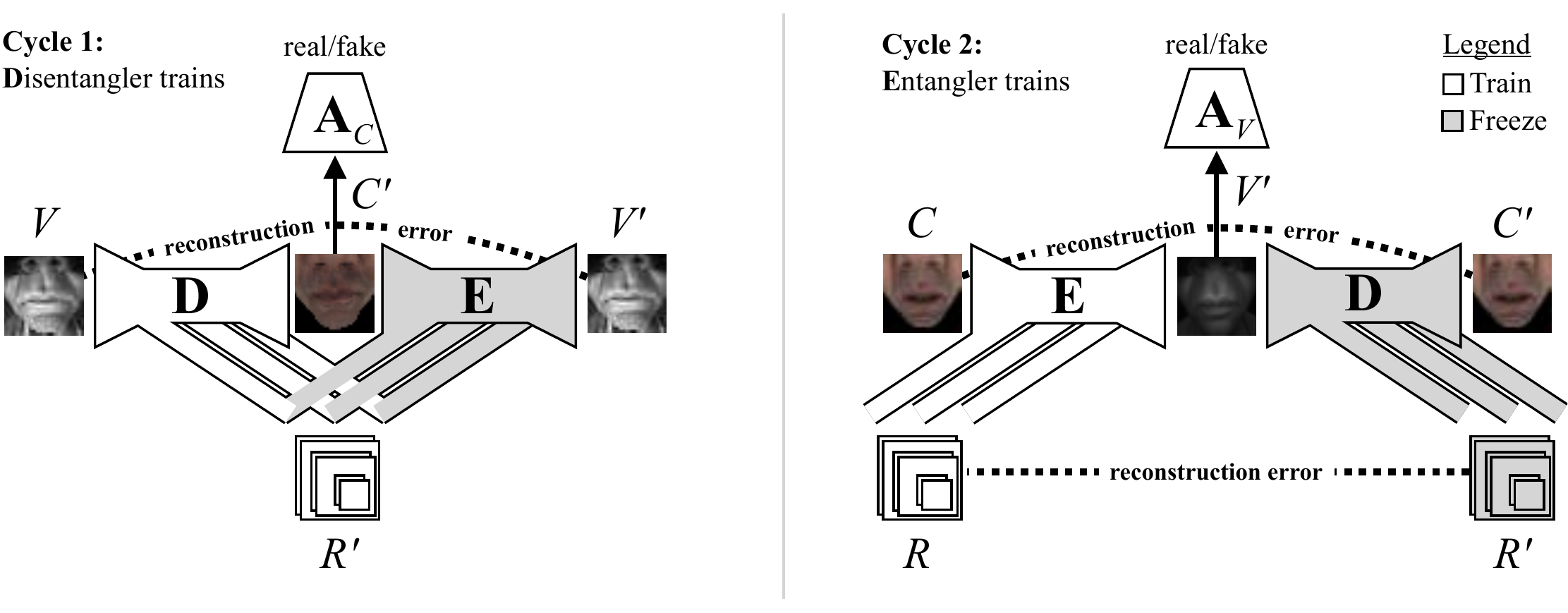}
\vspace{-1em}
\caption{Disentanglement-Entanglement architecture. The networks are $\bo{D}$ (disentangler), $\bo{E}$ (entangler), $\bo{A}_C$ (adversary on $C$), and $\bo{A}_V$ (adversary on $V$). Primes indicate tensors that are generated by the model during the forward pass. Each generated tensor is subject to either reconstruction loss or an adversarial loss. $R$ is a multi-scale output of $\bo{D}$; it is concatenated to the featuremaps of $\bo{E}$ at the corresponding scales. 
}
\label{fig:arch}
\vspace{-1.5em}
\end{figure*}

Our goal is to learn functions that can map between these domains. We call the first mapping a disentanglement, denoted $\D$, since it performs an intricate splitting operation: $\D(v) = (c,r)$. We call the second mapping an entanglement, denoted $\E$, since it performs a merging operation: $\E(c,r) = v$. 
Figure~\ref{fig:arch} relates the notation to the data and architecture. Note that $\D$ and $\E$ are inverses of one another. 

Our input is a set of samples $\{c_i\}_{i=1}^N$ from $C$, and samples $\{v_j\}_{j=1}^M$ from $V$. The two datasets are unpaired, and true correspondences might not exist. 

\subsection{Adversarial priors}

Our model has two main networks, $\D$ and $\E$. We would like to have $\D : \vcr$, and  $\E : \crv$. To achieve this, we introduce adversarial networks $\A_C$ and $\A_V$, which learn and impose priors on the distributions of our networks' outputs. 

The adversarial networks attempt to discriminate between real and fake (\ie, generated) samples of the domains $C$ and $V$. In our notation, we distinguish ``fake'' samples with a prime symbol. 
We train our main networks against the adversarial labels with the least-squares loss \cite{mao2017least}:
\eq{ \mathcal{L}_{\text{GAN}} =  \left[ \AV(v')-1\right]^2 + \left[\AC(c')-1 \right]^2 .}
In a separate (but concurrent) optimization, we also train the parameters of the adversaries, with the losses
$\mathcal{L}_{\AC} =  \left[\AC(c)-1\right]^2 + \left[\AC(c')\right]^2$, and $\mathcal{L}_{\AV} =  \left[\AV(v)-1\right]^2 + \left[\AV(v')\right]^2$. 

Note that we have no priors on the $r'$ samples generated by $\D$, because there is no dataset of ``true'' $r$ samples. Prior works manufactured a prior by assuming that $R$ is a low-dimensional Gaussian distribution (\eg, 8 dimensions, with zero mean and unit variance) \cite{huang2018multimodal,lee2018diverse,gonzalez2018image,liu2017unsupervised}. Here, we avoid this limiting assumption. We are able to do this because of our unique optimization procedure, detailed in Sec~\ref{sec:optim}. 
However, we do obtain some constraints on $R$ by enforcing cycle consistency, described next.


\subsection{Cycle consistency}

On each training step, the model runs two ``cycles''. Each cycle generates a reconstruction loss, which constrains the model to perform consistent forward-backward translation. Figure~\ref{fig:arch} illustrates the cycles. 

In the first cycle, the disentangler $\D$ receives a random $v$ from the dataset, and generates two outputs: $(c', r') = \D(v)$. These outputs are passed to the entangler $\E$, which generates $v' = \E(c',r')$. If the disentanglement and re-entanglement are successful, this output should correspond to the original $v$. Therefore, we form the reconstruction objective $\ell_v = \norm{v - v'}_1$, where $\norm{\cdot}_1$ denotes the L1 norm. In summary, this cycle performs $\E(\D(v)) = v' \approx v$.

The second cycle is symmetric to the first. The entangler $\E$ receives a random $c$ from the dataset, and an $r$ generated from a random $v$. Note that it is necessary to use generated $r$ samples here, since $R$ is completely determined by the network. We omit the prime on this $r$ since it is treated as an input rather than an output.
From the input $(c,r)$, the entangler generates $v' = \E(c,r)$. We then pass $v'$ to the disentangler, which generates two new outputs $(c', r') = \D(v')$. If the entanglement and disentanglement are successful, these outputs should correspond to the original inputs. We therefore form the reconstruction objectives $\ell_c = \norm{c - c'}_1$ and $\ell_r = \norm{r - r'}_1$. In summary, this cycle performs $\D(\E(c,r)) = (c',r') \approx (c,r)$. 

Collecting the reconstruction objectives, we have 
\eq{ \mathcal{L}_\text{recon} = \lambda_v \ell_v + \lambda_c \ell_c + \lambda_r \ell_r.}

Observe that there is no ``fidelity'' objective on the translated tensor of each cycle (\ie, $c'$ in Cycle 1, and $v'$ in Cycle 2); these tensors only have an adversarial loss. In other words, there is nothing in the design to force $c'$ to correspond to $v$, or $v'$ to correspond to $c$, other than the back-translation error. As we will show in experiments, this back-translation requirement is not sufficient, because the networks are able to cooperate on the back-translation: when $\E$ is the back-translator, it can compensate for errors made by $\D$, and vice versa. 

In practice, many of these ``errors'' are never corrected. Instead, they are adapted and refined, to minimize the adversarial loss while facilitating reconstruction. We call these ``cheats'': undesirable outputs that yield near-zero loss. 
At convergence, cheats often take the form of a within-domain transformation: this causes the adversary to not impose a loss (since the output is still in the correct domain), yet allows the second network to (jointly) learn how to undo the transformation.
These cheats are especially visible in experiments with faces, likely because humans are so sensitive to faces \cite{ekman1980face}. Figures~\ref{fig:teaser} and~\ref{fig:translate} show clear examples of this cheating behavior: while the two domains only differ in texture/lighting, the networks learn to additionally (and unpredictably) alter the facial expression. 

We observe that this undesirable solution to the reconstruction error requires both $\D$ and $\E$ to be complicit in the scheme. 
For example, if $\D$ transforms its input while translating it, but $\E$ is unaware of the cheat, $\E$ will not undo the transformation while back-translating, yielding a loss. 
This leads us to our optimization procedure, which essentially prevents $\D$ and $\E$ from cooperating in this way.










\subsection{Uncooperative optimization}\label{sec:optim}

The total loss we wish to minimize is 
${
\mathcal{L}_\text{total} = \mathcal{L}_\text{recon} + \mathcal{L}_\text{GAN}.
}$
As long as the forward translations land in the target domains, $\mathcal{L}_\text{GAN}$ is minimized; as long as the backward translations reconstruct the inputs, $\mathcal{L}_\text{recon}$ is minimized. 

As explained, there is a local minimum to this loss, in which forward translation includes an undesirable transformation, and back-translation includes an inverse transformation. This ruins the fidelity of the translation. 

To reach this bad minimum, each network needs to learn \textit{two} functions: (1) translating ``real'' inputs into outputs in a target domain, and (2) decoding ``fake'' inputs by undoing generated translations. Referring to Figure~\ref{fig:arch}, the disentangler $\D$ learns the first task in Cycle 1, and learns the second task in Cycle 2; the entangler $\E$ learns these functions in the opposite order. 
To prevent this from happening, we prevent the networks from learning how to decode fake inputs. We do this by \textit{freezing} the networks when they receive fake inputs. When a network is ``frozen'', it is treated as a fixed but differentiable function, so that gradients flow through it, but it does not learn. Referring again to Figure~\ref{fig:arch}, this means training $\D$ only in the first cycle, and training $\E$ only in the second cycle (where they respectively receive real inputs). 


With this optimization technique, the networks are incapable of learning how to compensate for each other's errors. This means that an erroneous forward translation will always be taken ``at face value'' by the backward translator, and produce an appropriate loss. This is because the backward translator's only experience (in terms of gradient steps) comes from real data. 

This method is a type of alternating optimization, in the sense that we keep one set of parameters fixed while optimizing the other set, and alternate. In practice, we alternate on every step. Specifically, we do a forward pass through Cycle 1, freeze $\E$ while we update $\D$, and then do a forward pass through Cycle 2, and freeze $\D$ while we update $\E$. 

Including the independent update required for the adversarial networks, this setup requires three optimizers in total.




\subsection{Implementation details}\label{sec:impl}

\paragraph{Network architecture} Our implementation is based on CycleGAN \cite{zhu2017unpaired}. The translators' architecture originally comes from Johnson \etal \cite{johnson2016perceptual}: two stride-2 convolutions, four residual blocks, and two transposed convolutions. 

We implement the disentangler as two separate networks: one for the $C$ stream and one for the $R$ stream; the $R$ stream ends before the transposed convolutions. We found this worked significantly better than using a single network to produce both $C$ and $R$.

The entangler uses the same architecture, except it receives skip connections from the $R$ stream. There are three such connections: the first uses the featuremap produced after the stride-2 convolutions; the second uses the featuremap after two residual blocks, and the third uses the featuremap after the next (and final) two residual blocks. These $R$ featuremaps are simply concatenated with the corresponding featuremaps in $\E$. The intent with multiple skip connections is to allow the network the capacity to transfer residuals at multiple levels of scale and abstraction. Our model has fewer residual blocks than CycleGAN, but the added $R$ stream makes the total parameter count similar. 

For discriminators, we use the $70 \times 70$ PatchGANs \cite{isola2017image} which were also used in CycleGAN. In all models, we apply spectral normalization to the weights of the discriminators \cite{miyato2018spectral}, which we found to stabilize the adversarial training. 

\paragraph{Training} We set the reconstruction coefficients on $V$ and $C$ to be ten times the GAN loss, so $\lambda_v = \lambda_c = 10$. We use a smaller coefficient on the the $R$ reconstruction, since it is a much larger tensor: $\lambda_r = 0.1$. We update the discriminator using generated images drawn randomly from a history buffer of size 50. We use the Adam solver \cite{kingma2014adam}, with $\beta_1 = 0.5, \beta_2 = 0.999$, a batch size of 4, and a learning rate of 0.0002. After the reconstruction errors stop descending, we linearly decay the learning rate to zero. In total, training can take up to 300,000 steps, which is approximately 3 days on a single Nvidia GTX 1080 TI. This is slower convergence than a traditional CycleGAN (which takes ~100,000 iterations on our data), likely because the objective is harder to optimize when ``cheating'' is disallowed.


\paragraph{Simplified settings for synthetic data} For the experiments with synthetic data, we use a model with fewer parameters. We implement each generator as a fully-connected network with one hidden layer of 32 units and ReLU activation. We implement each adversarial discriminator as a fully-connected network with one hidden layer of 32 units, and leaky ReLU activation. Our experiments suggest that the discriminators have more than sufficient capacity to correctly learn the distributions of $C$ and $V$ and keep equilibrium with the generators. We use the same training setup as in the real-image experiments, except we set the batch size to 128, and training to convergence takes approximately 60,000 iterations, which is 1 hour on a single GPU. 
\begin{figure}[t!]
\centering
  \begin{minipage}[b]{0.5\linewidth}
    \includegraphics[width=\textwidth]{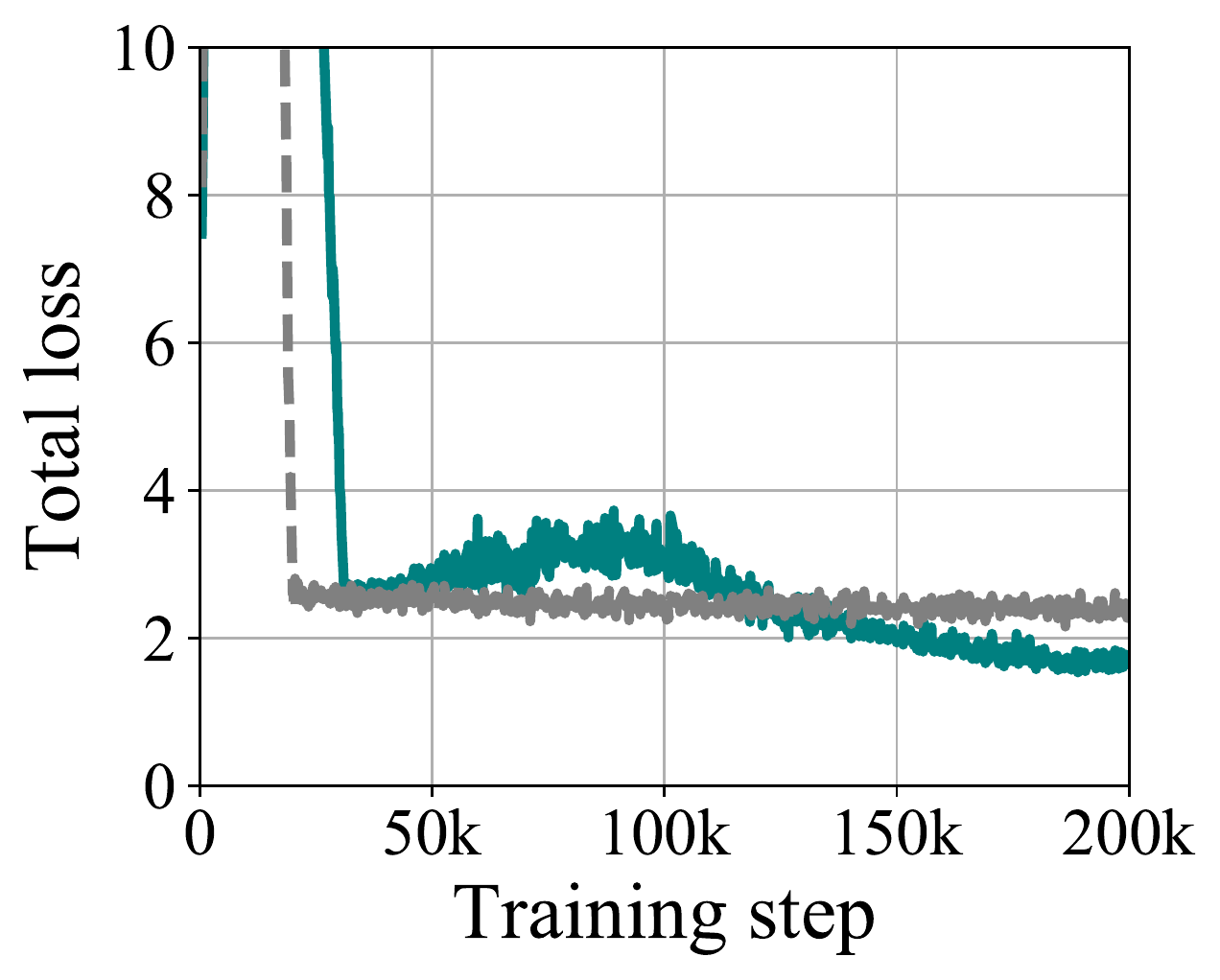}
  \end{minipage}\begin{minipage}[b]{0.5\linewidth}
   \quad
    \includegraphics[width=\linewidth]{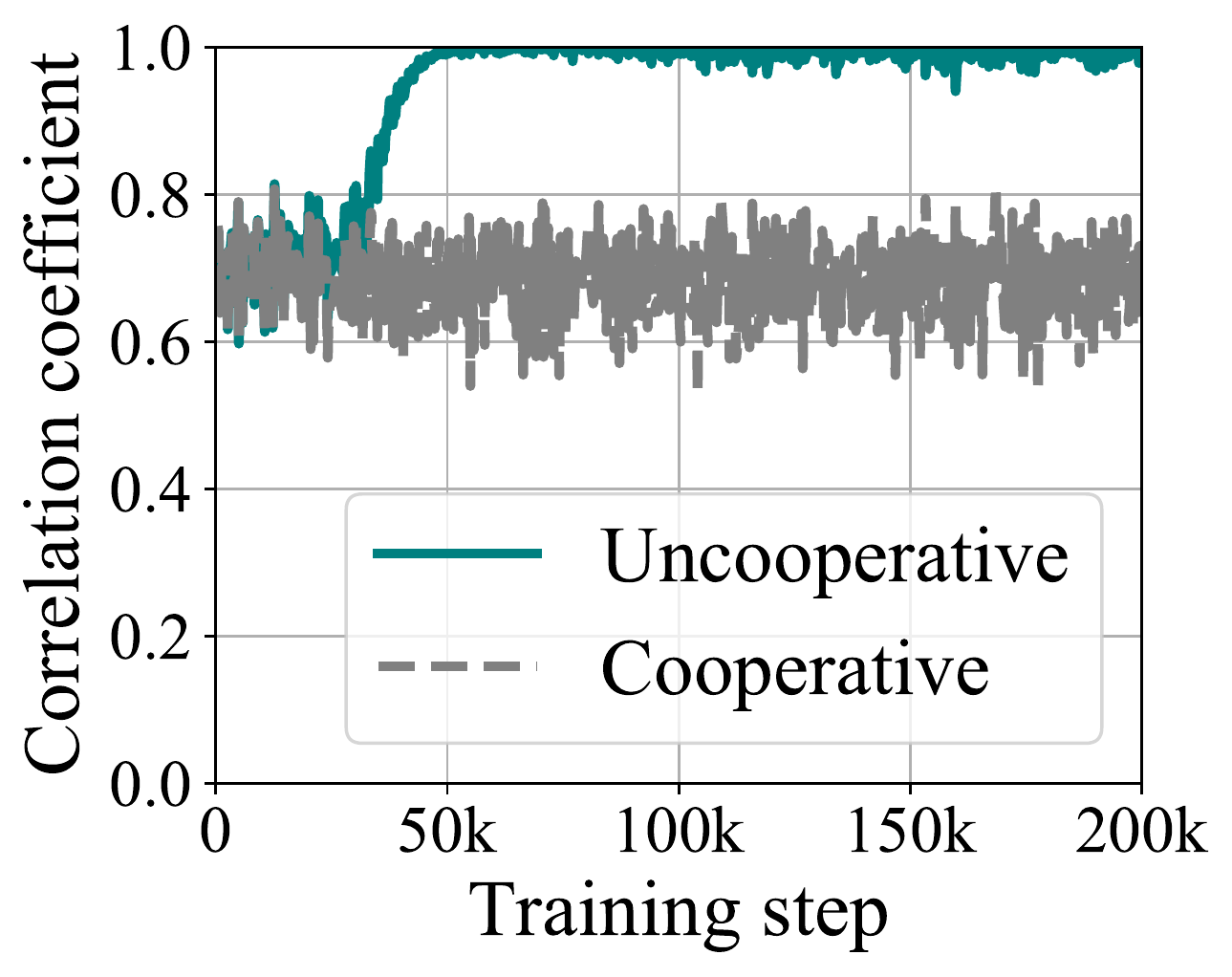}
  \end{minipage}
  \vspace{-2em}
  \caption{Uncooperative \vs cooperative optimization results on the objective function (left) and correlation with the ground-truth latent factor (right), over training steps. At convergence, uncooperative optimization achieves near-perfect disentanglement of the latent factor, whereas cooperative optimization does not. }
    \label{fig:plots}
    \vspace{-1em}
\end{figure}

\section{Experiments} \label{sec:experiments}

In this section, we demonstrate that our method outperforms prior work on (1) \textbf{accuracy} of disentanglement, (2) \textbf{fidelity} of translation, (3) \textbf{coverage} of modes (in multi-modal translations).
Ground-truth disentanglements do not exist in real image data, so we use a simple synthetic scenario to quantitatively evaluate accuracy, then present real-world qualitative results for fidelity and coverage. 

\subsection{Disentanglement accuracy}

One of our claims is that uncooperative optimization is critical for accurate disentanglement. This is based on the idea that a uncooperative models are less able to find ``cheats'' which bypass the need for accuracy.

In other words, we need to show that ``uncooperative'' optimization leads to correctly disentangling $R,C$ from within $V$, in a setting where ``cooperative'' optimization fails. 
We present one in which the ground-truth factors are 1D, and entanglement/disentanglement is simply concatenation/splitting. We find that cooperative optimization is incapable of learning this simple operation, whereas uncooperative optimization succeeds. 


\paragraph{Models} In this experiment, we use two \textit{identical} models (see the ``Simplified settings for synthetic data'' in Sec.~\ref{sec:impl}), and change only the optimization method: one uses the proposed ``uncooperative'' optimization, and the other uses the baseline ``cooperative'' optimization.  

\paragraph{Data} Since ground-truth latent factors are generally unknown in real data, it is necessary to design synthetic data for this experiment. 
We define the latent factors $C$ and $R$ to be Gaussian distributions. We generate synthetic entanglements $v_i \in V$ by concatenating a sample $c_i \sim C$ with a sample $r_i \sim R$. 
Specifically, we draw the elements of $C$ from a 1D Gaussian with $\mu=2.0, \sigma=1.0$, and draw the elements of $R$ from a 1D Gaussian with $\mu=-2.0, \sigma=1.0$. We find that results are not sensitive to dimensionality (except in convergence time), and so present only the simplest version here, setting the dimensionality of both $C$ and $R$ to 1, making the dimensionality of $V$ equal to 2. Note that the $R$ domain is never encountered at training time, except in its entangled form inside $V$. The task is to recover $R$, using only disentanglement/entanglement cycles, and unpaired samples of $V$ and $C$.

\paragraph{Metrics}

We measure the similarity of the actual $R$ domain (used to generate $V$ samples) to the learned $R'$ domain (disentangled from $V$ samples) using the absolute value of the Pearson correlation coefficient $\rho = |\text{cov}(R,R')/(\sigma_{R} \sigma_{R'})|$, which equals 1 if the two variables have a totally linear relationship, and is closer to 0 otherwise. This (unlike a distance metric) allows solutions where the learned $R'$ is a scaled version of the true $R$, which is appropriate since scaling may be absorbed in the model weights. 




\begin{figure}[t!]
\centering
\includegraphics[width=\linewidth]{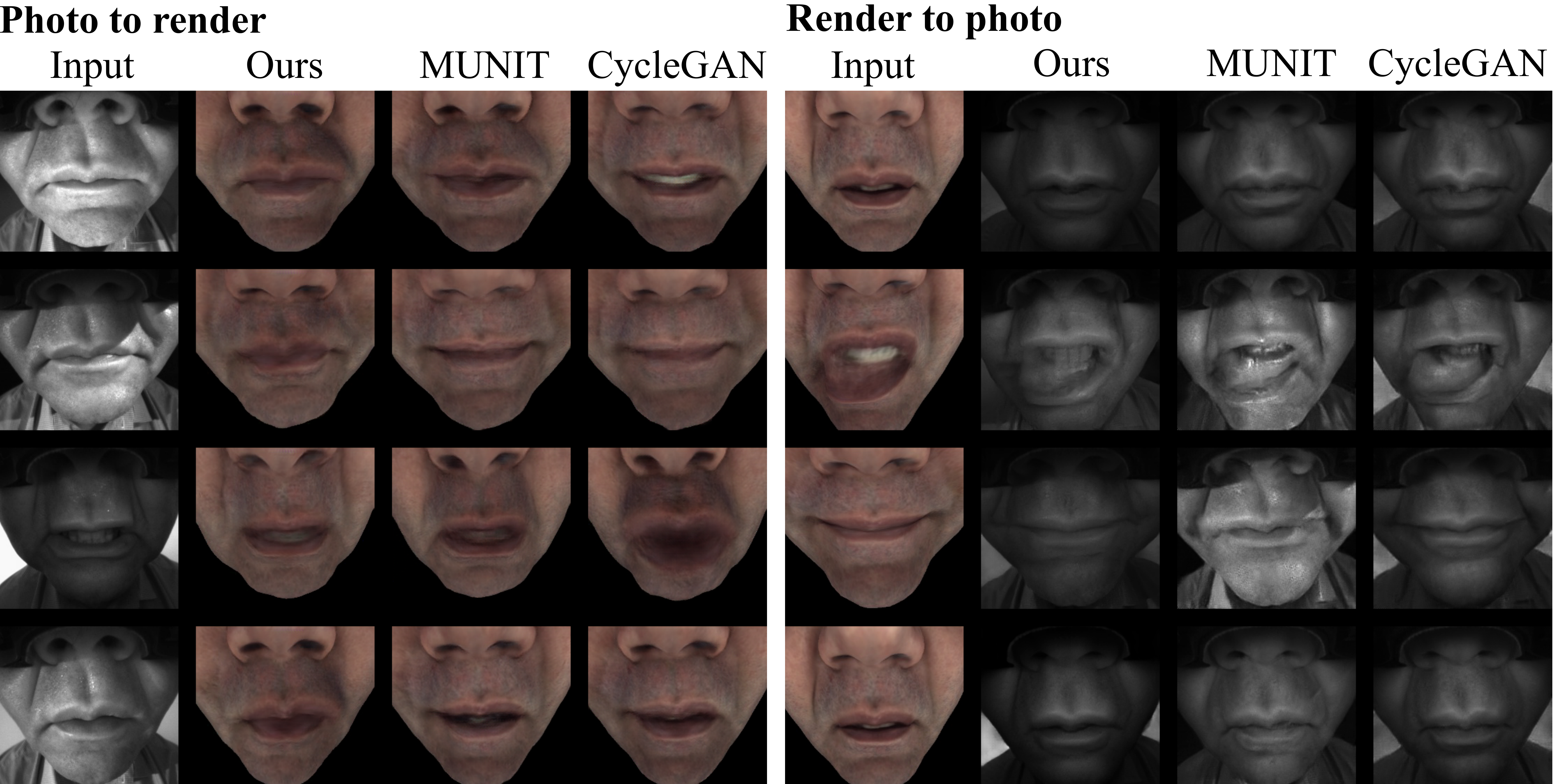}
\caption{
Domain translation results on the face dataset, compared with MUNIT and CycleGAN. In both translation directions, MUNIT and CycleGAN sometimes alter the expression of the subject; our method keeps expression intact.
}
\label{fig:translate}
\vspace{-1em}
\end{figure}

\paragraph{Results} Our results are summarized in Figure~\ref{fig:plots}. The two models converge in approximately the same number of iterations. At the end of training, the cooperative version achieves a correlation coefficient of 0.695, while the uncooperative version achieves 0.998. Results vary slightly across iterations (and across initializations), but correlation does not noticeably improve for the cooperative version, even if training is extended to 200k iterations. 

Overall, this shows that uncooperative optimization leads the model to disentangle the \textit{true} latent factors, while cooperative optimization does not. 



\subsection{High-fidelity translation}

One of our claims is that the uncooperative training leads to \textit{high-fidelity} translations. By this, we mean that the translation retains as much information as possible from the input, without altering it. To evaluate this, we compare our compare our model's \textit{forward translations} against those of CycleGAN and MUNIT. 

\paragraph{Baselines}
CycleGAN is a popular baseline in unsupervised (but unimodal) image-to-image translation; our architecture is based on it. MUNIT is a state-of-the-art unsupervised \textit{multimodal} image-to-image translation method.

\paragraph{Data} We note that MUNIT was originally applied to translating between widely different domains, \eg, translating dogs to lions. While this type of translation is impressive, it is also difficult to evaluate, and it is not clear that close pixel-wise correspondence/fidelity is even desirable in that task.

In this paper, we primarily focus on translating a human face across two appearance domains: photos of the face captured by a head-mounted camera, and renders of the face produced by a parametric face model (already adapted to the input face). This has an application in social virtual/augmented reality (VR/AR), where we would like users to interact with each other ``face-to-face'' (inside the virtual environment) as naturally as possible. 

We collected the face data ourselves. 
The real photos (representing the $V$ domain) were captured by a camera attached to the actor's headset, with the lens pointed toward the bottom half of the actor's face; lighting variation was achieved with a set of lights surrounding the actor; background variation was achieved by placing large computer monitors behind the actor and displaying random images. Rendered images of the same face (representing the $C$ domain) were produced by fitting a deep parametric face model to the actor \cite{lombardi2018deep}, and generating random expressions from a viewpoint similar to the headset view. There are 7074 real photos, and 1000 rendered images. 
The task is to translate a photo of a face to (or from) a rendered-like image of the same face, while maintaining the face's expression.  

For completeness, we also show results on translating architectural facades $\leftrightarrow$ labels \cite{tylevcek2013spatial}, which is a task used in prior work \cite{zhu2017unpaired}. We have also experimented with the aerial photos $\leftrightarrow$ Google maps task \cite{zhu2017unpaired}, but did not find noticeable differences between the methods on that task.
 

\begin{figure}[t!]
\centering
\includegraphics[width=1.0\linewidth]{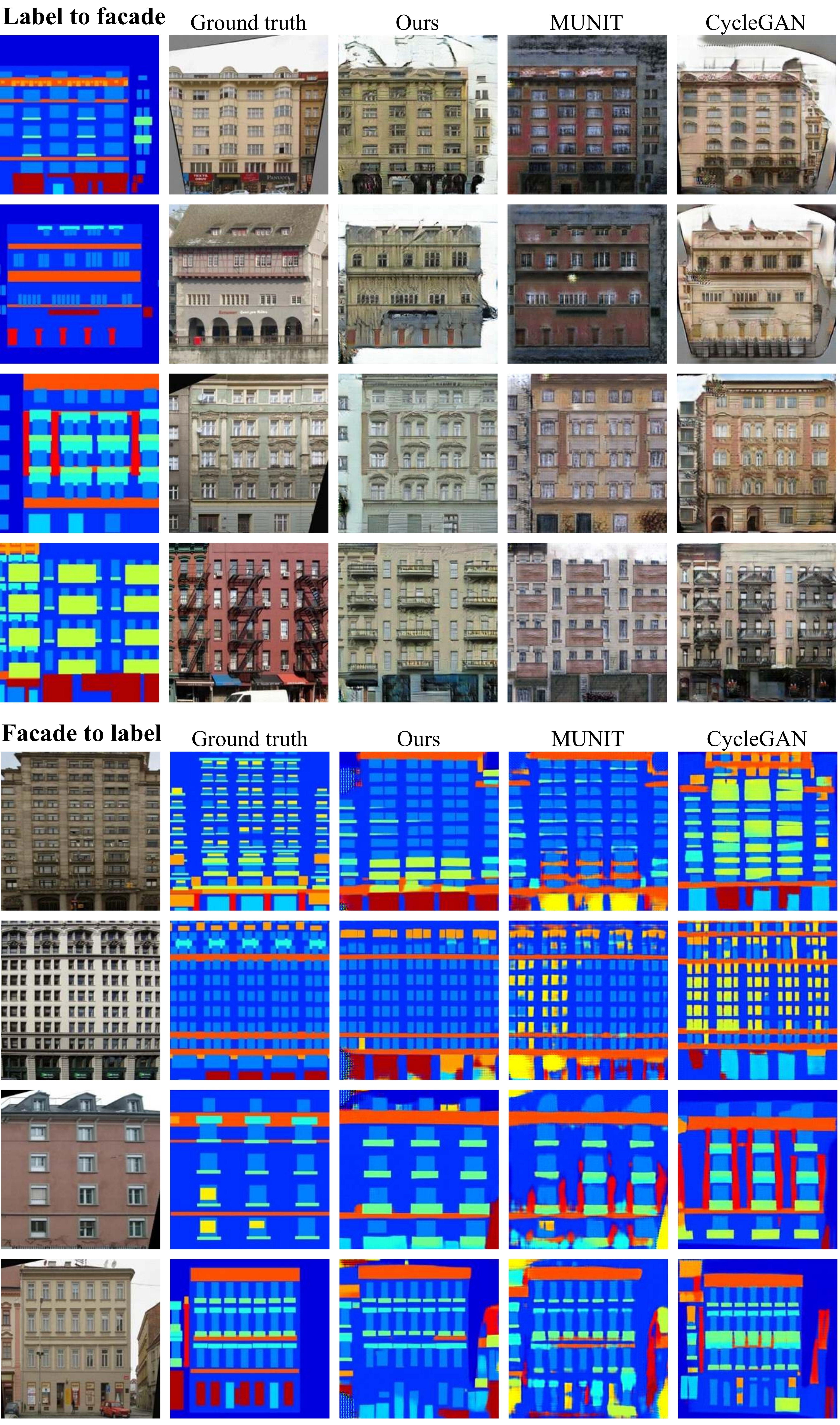}
\caption{
Domain translation results on facade/label images. While MUNIT and CycleGAN introduce artifacts (which make back-translation easier during training), our model performs high-fidelity translation.
}
\label{fig:facades}
\vspace{-1.2em}
\end{figure}

\paragraph{Metrics}
In the face image experiments---which are necessarily qualitative---we rely on the fact that humans are extremely adept at reading faces \cite{ekman1980face}, and attempt to demonstrate that our model achieves obviously better disentanglements than prior methods. The results on aerial and facade data (introduced in prior work) is harder to interpret at a glance, but close inspection can reveal differences in sharpness and  spatial consistency with the input. We note that even when ground truth translations exist, it does not make sense to evaluate against them, since these are many-to-one/one-to-many mappings, and totally unsupervised models (as considered here) cannot be expected to generate labels that match the ground truth (\eg, as assumed in the ``FCN score'' used in Pix2Pix \cite{isola2017image} and CycleGAN \cite{zhu2017unpaired}). 
\paragraph{Results}
Figure~\ref{fig:translate} compares our method against MUNIT and CycleGAN on the face dataset. 
The results show that while CycleGAN and MUNIT perform the appearance translation, they make small but very noticeable shifts in the facial expression, \eg, turning a closed mouth into a smile, or changing a grimace to a pout. This is due to the drawbacks of cooperative training, described earlier. Our method does not have this problem, and translates the faces across domains without altering expression. Figure~\ref{fig:facades} shows the same experiment but for the facades $\leftrightarrow$ labels task, with similar results: while our method retains, for instance, exact spatial positions of the features in either domain, the baseline methods tend to make small shifts in position and scale.



\subsection{Multi-modal outputs}

Our model is designed to produce multi-modal outputs, through a ``mix-and-match'' method, where we use $C$ from one input and $R$ from another input, and entangle these to form a novel sample of $V$. We compare against MUNIT, which is the current state-of-the-art method for this task. 

More specifically, generating multiple outputs from a single input involves the following steps: (1) given $v_i$ as input, generate $c_i$; (2) given an unrelated $v_j$ as input, generate $r_j$; (3) entangle $c_i, r_j$, to produce the composite $v_{ij}$. In the face context, since the domain $C$ contains expression but not lighting, this setup means extracting expression from one image, and extracting everything else (which is mostly lighting and backgrounds) from another image, and combining these factors into a new image. The experimental setup is similar for MUNIT: a ``content code'' is generated from $v_i$, and a ``style code'' is generated from $v_j$, and these are encoded into the final output. We do this for multiple $v_j$, to show the effect of transferring a variety of residual factors onto the same face. 


\paragraph{Data} We use the same face data as in the high-fidelity task, and also aerial photos $\leftrightarrow$ Google maps \cite{zhu2017unpaired}, which we find has more evident multi-modality than the facades. 



\begin{figure}[t]
\centering
\includegraphics[width=\linewidth]{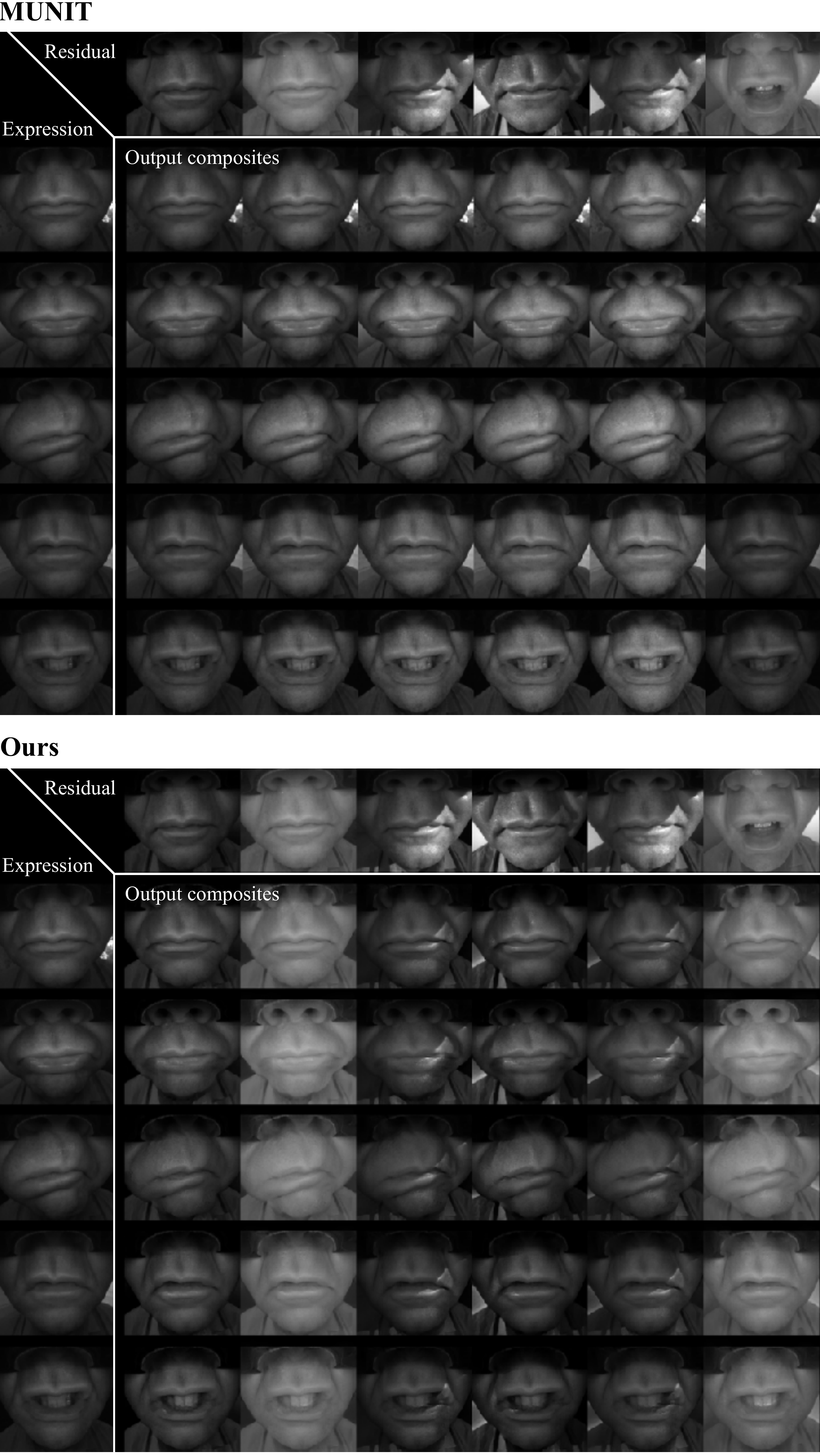}
\caption{
Face relighting results of MUNIT (top) vs our model (bottom). In each table, the leftmost column shows the input $c_i$ from which expression is drawn; the top row shows the input $v_i$ from which everything else is drawn. 
}
\label{fig:relighting}
\vspace{-1.5em}
\end{figure}

\paragraph{Results} Figure~\ref{fig:relighting} shows the results of this experiment on the faces dataset, for MUNIT and our model. 
The figure shows expressions from $v_i$ across rows, and residuals from $v_j$ (\ie, lighting/background conditions) across columns. 
For an overview of the results, the reader may scan across rows to inspect that expression is transferred from the leftmost row, and scan across columns to inspect that lighting and backgrounds are transferred from the topmost row.
MUNIT appears to have only learned to transfer the global intensity from the $v_j$ source. Our model appears to be transferring backgrounds, and even casting distinct shadows onto the face. However, some shadows appear reduced in intensity (\eg, third column), suggesting that expression-lighting disentanglement is not perfect here. 

In the supplementary, we also show results of this experiment on the aerial photos $\leftrightarrow$ Google maps dataset, where we treat the Google map as $C$ (assuming it has less information), and the aerial photos as $V$. 
In this domain, it appears MUNIT transfers very little from the residual, while our model incorporates textures and objects (\eg, note the white object transferred from the first residual). 
Both methods appear to retain the spatial layout of the input map.


\section{Discussion}
\label{sec:conclusion}

In this work, we address the \textit{compensation} issue in translation cycle-consistency, which typically diminishes the utility of the reconstruction loss. In compensation, the back-translator (undesirably) adapts to the weaknesses and shortcuts of the forward-translator. Hypothetically, there is another way to (partially) defeat the loss, which may be called \textit{exploitation}. 
In exploitation, the forward-translator (undesirably) adapts to the weaknesses and shortcuts of the back-translator. The enduring exploitation issue may explain the subtle imperfections in our outputs. 

Another limitation of our approach is that we do not address \textit{many-to-many} mappings. Our approach is only multi-modal in one direction. 

In summary, we introduced the problem of high-fidelity image-to-image translation, motivated it for augmented reality applications, and presented an unsupervised method for solving it. 
We identified a fundamental cause of low-fidelity translations: cooperation between the forward translator and the backward translator, which allows the forward-translation to ``hide'' information, and the back-translator to ``recover'' from noticeable errors. This is a critical problem in real applications. We presented an ``uncooperative'' optimization scheme that prevents the problem. Our results demonstrate that uncooperative optimization leads to high-fidelity image translations, making image-to-image translation not only fun, but useful for augmented reality.



\clearpage
{\Large \bf \centering \noindent Supplementary Material}

\setcounter{section}{0}
\renewcommand{\thesection}{\Alph{section}}

\section{How ``cheating'' happens in practice}

It is relatively easy to see how the ``uncooperative'' optimization prevents the networks from developing a ``cheating'' scheme, since the networks only train when their inputs are real. It is less easy to see how a ``cheating'' scheme can develop at all, considering the losses that already constrain the model. 
In this section, we will first summarize a tempting (but flawed) argument suggesting that ``cheating is penalized by the losses'', and then demonstrate how the intuition is generally proven wrong in practice.

To see how cheating may intuitively seem impossible, consider the following, with reference to Figure~3 in the main text. 
Suppose $r$ is used as a ``shortcut'' to cheat Cycle 1, in the sense that $\D$ copies $v$ into $r'$, and then $\E$ copies $r'$ into $v'$, meeting the cycle-consistency constraint of $v' \approx v$. Meanwhile, to meet the adversarial constraint, $\D$ may write any target-domain image into $c'$.
But this leads to errors in Cycle 2: if $\E$ simply copies its input $r$ into $v'$, and/or $\D$ does not produce an output $c'$ which strictly corresponds to its input $v'$, then $c$ is essentially ignored, and we will have $c' \neq c$ and a loss. Therefore, it seems that cheating should be eliminated at convergence. 

\begin{figure}[t!]
\centering
\includegraphics[width=\linewidth]{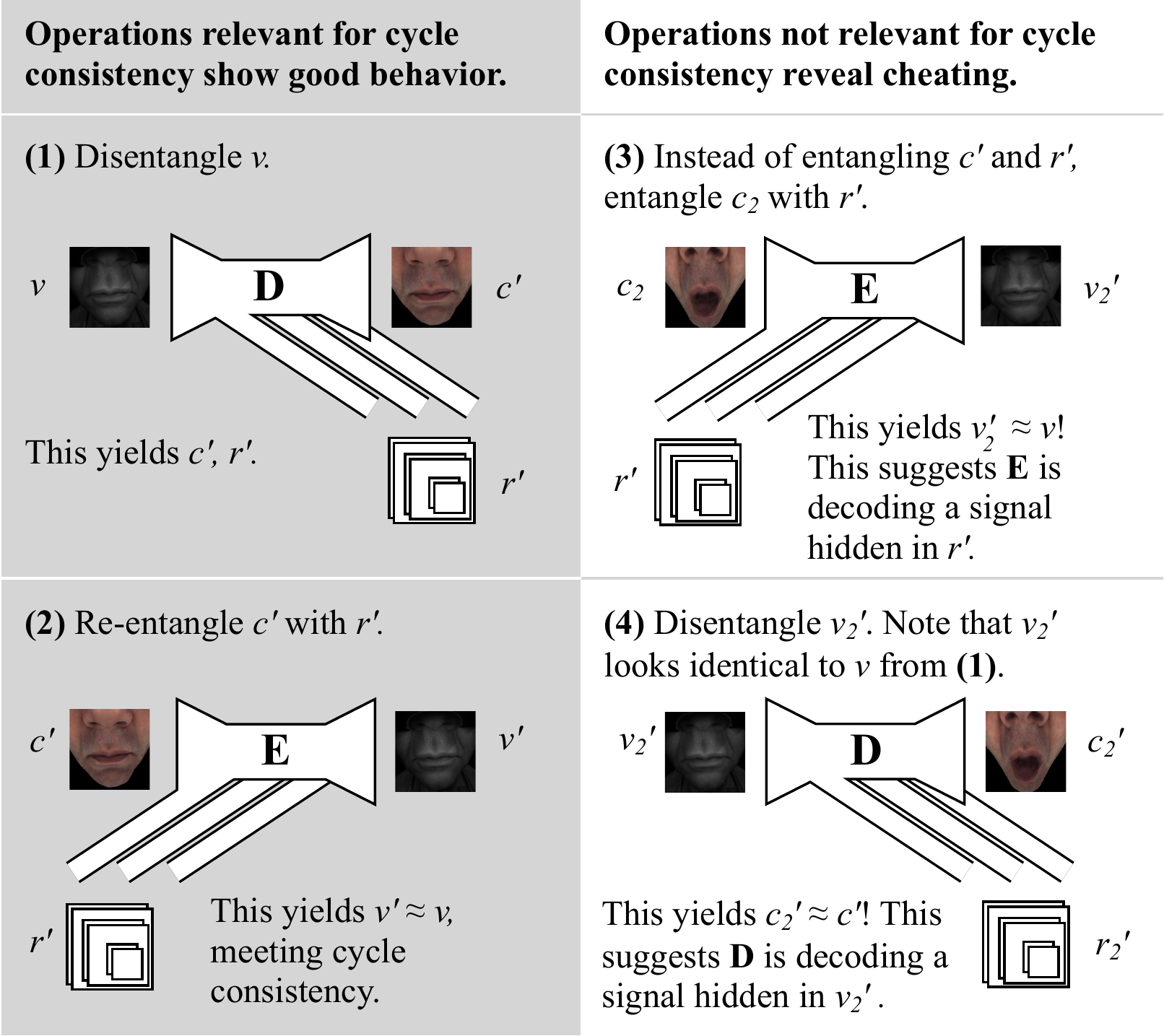}
\caption{{Exploration of cheating behavior.} 
    Left: Decoding \textit{matched} tensors yields apparently good output. Right: decoding \textit{mismatched} tensors reveals clear cheating. 
    }\label{fig:cheat}
\end{figure}

In practice, however, the networks achieve a far more subtle type of cheat, which eventually yields zero loss. At training time, the visual manifestation of the cheat is that the translations do not correspond to the inputs, and yet they are back-translated perfectly. 
Our experiments suggest that the networks generate outputs that facilitate reconstruction of the corresponding inputs, and the networks treat these generated tensors differently from real tensors. 
In particular, when we generate $(c', r') = \D(v)$, then $r'$ tends to hide $v$ inside, to facilitate its reconstruction by $\E$. Similarly, when we generate $v' = \E(c,r)$, then $v'$ tends to hide $c$ inside, to facilitate its reconstruction by $\D$. 
Figure~\ref{fig:cheat} illustrates how to empirically reveal this behavior, and shows sample non-corresponding outputs from a converged ``cooperative'' model. For a brief reading of the figure, 
observe that $v$ and $v_2'$ appear visually identical, but $\D$ decodes (the real) $v$ into a closed mouth, and decodes (the fake) $v_2'$ into a wide open mouth. 





Parallel work \cite{chu2017cyclegan} has also observed this phenomenon, under the label of \textit{steganography}. That work showed that the secret/cheating signal is often hidden in high frequencies, where presumably the discriminators are less effective. With sufficient training, a discriminator should learn to block this strategy (since such high-frequency content is not present in real examples), which would force the signal to shift to lower (and more semantically-relevant) frequencies, as observed here.

\section{Additional results}
\begin{figure*}[t]
\centering
\includegraphics[width=\linewidth]{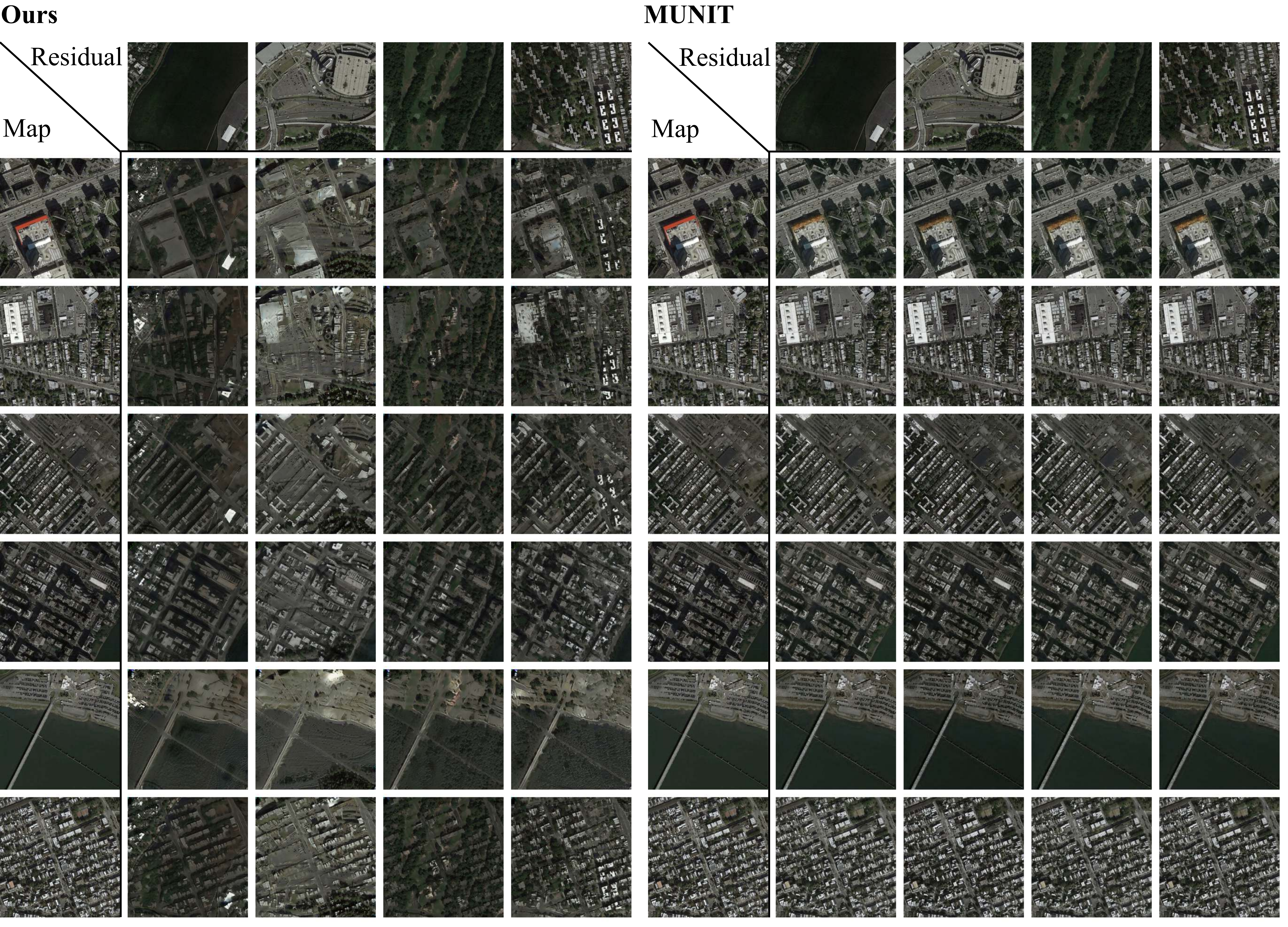}
\caption{
Aerial image composition of our method (left) vs MUNIT (right). Our method successfully transfers textures from the residual, while MUNIT does not; both retain the spatial structure of the map in this case.
}
\label{fig:aerial_relighting}
\end{figure*}

Figure~\ref{fig:aerial_relighting} shows results on the aerial photos $\leftrightarrow$ Google maps dataset. 
Both methods appear to retain the spatial layout of the input map, but our method transfers textures and lighting/brightness from the residual, while MUNIT does not.

\end{document}